\documentclass{article}

\usepackage{arxiv}

\usepackage[utf8]{inputenc} 
\usepackage[T1]{fontenc}    
\usepackage{hyperref}       
\usepackage{url}            
\usepackage{booktabs}       
\usepackage{amsfonts}       
\usepackage{nicefrac}       
\usepackage{microtype}      
\usepackage{lipsum}		
\usepackage{graphicx}
\usepackage{natbib}
\usepackage{doi}

\usepackage{amsmath}
\usepackage{subcaption}
\usepackage{multirow}
\usepackage{mathtools}

\title{Increasing faithfulness in human-human dialog summarization with Spoken Language Understanding tasks}

\author{
Eunice Akani \textsuperscript{1,2},
Benoit Favre \textsuperscript{1},
Frederic Bechet \textsuperscript{1,3},
Romain Gemignani \textsuperscript{2}\\
\textsuperscript{1} Aix-Marseille Univ, CNRS, LIS, Marseille, France \\ 
\textsuperscript{2} Enedis, Marseille, France\\
\textsuperscript{3} International Laboratory on Learning Systems - ILLS - IRL CNRS, Montréal, Canada\\
\texttt{eunice.akani@univ-amu.fr}, 
\texttt{firstname.lastname@lis-lab.fr}, 
\texttt{romain.gemignani@enedis.fr}
}

\date{}

\hypersetup{
pdftitle={Increasing faithfulness in human-human dialog summarization with Spoken Language Understanding tasks},
pdfsubject={sc.CL, sc.AI},
pdfauthor={Eunice Akani, Benoit Favre, Frederic Bechet, Romain Gemignani},
}

\begin{document}
\maketitle

\begin{abstract}
Dialogue summarization aims to provide a concise and coherent summary of conversations between multiple speakers. While recent advancements in language models have enhanced this process, summarizing dialogues accurately and faithfully remains challenging due to the need to understand speaker interactions and capture relevant information.
Indeed, abstractive models used for dialog summarization may generate summaries that contain inconsistencies. 
We suggest using the semantic information proposed for performing Spoken Language Understanding (SLU) in human-machine dialogue systems for goal-oriented human-human dialogues to obtain a more semantically faithful summary regarding the task.
This study introduces three key contributions: First, we propose an exploration of how incorporating task-related information can enhance the summarization process, leading to more semantically accurate summaries. Then, we introduce a new evaluation criterion based on task semantics. Finally, we propose a new dataset version with increased annotated data standardized for research on task-oriented dialogue summarization. 
The study evaluates these methods using the DECODA corpus, a collection of French spoken dialogues from a call center. Results show that integrating models with task-related information improves summary accuracy, even with varying word error rates.
\end{abstract}

\section{Introduction}
Advancements in spoken language processing, made possible by low word-error-rate automatic speech transcription and generative language models, have enhanced the quality of outputs across various tasks, including dialog summarization. Dialog summarization involves generating a concise, abstract rendition of the dialogue exchanged among speakers in a multi-party spoken interaction. Due to discrepancies between transcript and summary styles, inaccuracies in the transcript, verbal hesitations by participants, and the generative process itself, dialog summarization is susceptible to "hallucinations", wherein generative models produce text containing nonfactual information when compared to the source. 

A study by \cite{cao-etal-2018-FaithfulTT} found that 30\% of summaries generated by text summarization systems contained incorrect information, known as "hallucinations" \cite{maynez-etal-2020-faithfulness}. Approaches to assessing summary faithfulness include textual entailment \cite{falke-etal-2019-ranking, maynez-etal-2020-faithfulness}, entity analysis \cite{nan-etal-2021-entity, ziwei-etal-2022-survey}, and question-answer verification \cite{durmus-etal-2020-feqa}.
Dialog summarization is growing in popularity but is challenged by dialog structure and diverse data types, such as customer service and technical discussions. Methods using auxiliary information like dialogue acts \cite{goo2018abstractive} or domain terminology \cite{koay-etal-2020-domain} have been proposed.
There are fewer studies on hallucinations in dialog summarization compared to text. \cite{wang-etal-2022-analyzing} found that 35\% of SAMSum dataset summaries were inconsistent with the source dialogues. \cite{tang-etal-2022-confit} identified eight types of factual errors in dialogue summarization, while \cite{wang-etal-2022-analyzing} identified six types and used a summary model based on conditional generation probabilities to evaluate the model's faithfulness.

In industrial applications, this issue may be more prevalent due to the use of smaller language models for cost, energy efficiency, and data protection. Leveraging domain knowledge can help evaluate faithfulness and improve summary accuracy.

In this paper, we aim to enhance dialog summarization faithfulness by integrating task-specific elements like caller intent and domain-specific named entities. Our study focuses on call centers, using the DECODA corpus \cite{bechet-etal-2012-DECODA}, a dataset of goal-oriented human interactions with semantic annotations. To our knowledge, no similar large, natural English speech dataset exists for this research. We also propose a new dataset using data augmentation via LLM to augment the dataset and improve the performance of dialog summarization model.

\section{Integration of task-related information in spoken dialog summarization}
On goal-oriented human-human conversations, we can leverage semantic information proposed for performing SLU in human-machine dialogue systems for tasks such as transportation reservations \citep{hemphill-etal-1990-atis} or restaurants/hotels \citep{devillers-etal-2004-french}. Typically, here, three semantic levels are defined: domain (the semantic context of the dialogue, such as flight bookings), intent (the nature of the request made in the dialogue, like confirmation or information inquiry), and entity/value pairs (the semantic relationships within intents, such as a destination in an itinerary request). 
As we focus on goal-oriented, we put ourselves in a customer service conversation. Thus, the call type can be seen as an intent in SLU and the domain named entities as a slot. 
We hypothesize that by incorporating task-specific semantic information into the summarization process, we can effectively address inconsistencies related to this information in the data. 
In this section, we will focus on the call type and entity level. Using call type and named entities is an excellent choice for our study as it is easily obtained and leveraging named entities provides a practical framework for our methodology. Nevertheless, other semantic information, such as semantic frames can be used.

\subsection{Summarization based on task information }
Identifying different call types and concept labels is essential for analyzing conversation recordings or transcripts, but this information often needs to be inferred. This can be done by implementing a separate prediction system on the transcript and integrating call types into the input data of the summarization system. The method is $\textbf{Pipeline}_C$: Predict semantic information $C$ from dialog $D$ such as $C=\textrm{intent}(D)$, then condition summary generation on $C$: $S = \textrm{summary}(D, \textrm{semantic-information}(D))$

\subsection{Task-related information as a selection criteria}
\label{subsec:summarySelection}
We propose using semantic information to select the most reliable summary from different summaries obtained by tweaking decoding parameters such as top-p. The method is based on call type prediction and the hallucination risk on entities.
\paragraph{Call type}
Based on the hypothesis that the generated summary should have the same call type as the dialog transcription, we suggest using call type as a summary selection criterion. Our method involves using a text classifier to predict the call type of the generated summary and comparing it to the predicted call type for the entire transcript. We handle multiple call types and uncertainty in call type classification by calculating the divergence between the probability distribution on all call types for both summary and dialog classifiers using the KL divergence \citep{kullback_information_1951}.
For $G$ and $R$ the probability distribution given by the call type classifier on the generated summary and the entire dialog respectively, $CT$ the set of call types; the KL divergence between $G$ and $R$ is defined as follows:

\begin{equation}
    D_{\text{KL}}(G \parallel R) = \sum_{x \in CT} G(x) \log \left(\frac{G(x)}{R(x)}\right)
\end{equation}

This allows us to choose the summary that minimizes the $D_{\text{KL}}$ among the generated summaries.

\paragraph{Named entities}
We propose using the NEHR criterion \citep{akani-etal-2023-reducing}, which measures the proportion of named entities not present in the source document, as a selection criterion for creating faithful summaries. This approach has shown favorable results in reducing hallucinated entities in text summarization. Thus, we decided to apply the same experiment to dialog summarization.

\paragraph{Combining NEHR and $D_{\text{KL}}$} This study aims to create a summary that accurately captures the task's semantic information. We hypothesize that combining criteria like NEHR and $D_{KL}$ will yield a summary consistent with the source document. For $H$ the set of summaries sampled from the model, $m$ the minimum NEHR value, $V=\left\{x\in H| NEHR(x) = m\right\}$ the set of summaries with the minimum NEHR, $D_{\text{KL}}(x \parallel R)$ the KL divergence between a summary $x$ associated with the dialog transcription $R$, the final system will be.

\begin{equation}
    \hat{s} = \mathop{\mathrm{min}}_{x\in V} D_{\text{KL}}(x \parallel R)
\end{equation}

\subsection{Task-related information as an evaluation criteria}
We propose a metric CT-Acc, which measures the accuracy of the call type classifier used on the generated summary compared to the reference call type. The assumption is that the summary is more coherent with the call type classifier if the accuracy is higher.

We propose using a metric based on named entities to assess the coherence of generated summaries compared to reference summaries. The metric measures entity-level accuracy and aims to ensure that both summaries contain the same entities. We measure NE-P, NE-R, and NE-F1 metrics to represent precision, recall, and F1 between entities in the generated and reference summaries. Our approach differs from previous work \cite{nan-etal-2021-entity} as we include all types of entities, such as dates and numerals, which we believe are crucial in some scenarios.

\section{Experiments}

\subsection{The DECODA Corpus}
\paragraph{DECODA-v1-hum} \citep{bechet-etal-2012-DECODA} contains spoken conversations between agents of the Paris Transport Authority and users of Paris buses and metro lines. Each conversation has a manual text transcription and a short summary. The corpus covers various call types such as \textit{Itinerary}, includes entities from a domain ontology such as \textit{Transport, and Schedule}, and consists of three parts (DECODA-1, DECODA-2, DECODA-3) with annotated synopses available only in DECODA-1 and DECODA-3.
Since DECODA-1 summaries are written in a non-literary style and with little detail, we decided to keep only those in DECODA-3 for system development.
From the 500 dialogues of the DECODA-3, we kept 200 dialogues as an evaluation set (\textit{test}), 200 for fine-tuning the system (\textit{Hum.} for \textit{human} annotated data) and 100 as a validation set to adjust the parameters. As not all DECODA summaries are written in a literary style and are sometimes non-existent for certain conversations, we create a new dataset with more annotated data for task-oriented dialog summarization.

\paragraph{DECODA-v2-aug}
We leverage prompt-based LLM to create summaries for the remaining DECODA conversations. The methodology entails processing the manual transcription with an instruction-based Language Model (LM) in a few-shot learning mode, providing the system with a few examples of target summaries and instructing it to produce similar summaries.
To do so, we decide to use ChatGPT-3.5 from OpenAI\footnote{\url{https://openai.com/index/chatgpt/}}  as the model have good capacity to produce coherent and fluent French text and perform various tasks without extensive finetuning. 
In addition to the gold synopses from previous train dataset (\textit{Hum.}), we add the summaries produced by ChatGPT-3.5 on the manual transcriptions of DECODA-1 and DECODA-2. We add some pair (dialog-automatic summary) to the development dataset and keep the same test set to form the new dataset call \textit{Aug.} for \textit{augmented}.
Appendix \ref{sec:prompt} shows the prompt used to generate summaries.

\paragraph{DECODA-v2-auto}
Then, we  put ourselves in a scenario where manual transcription was unavailable and opted to use an ASR system for transcribing the dialog. We used WhisperX \citep{bain2022whisperx}, a state-of-the-art model for ASR based on Whisper \citep{radford2022robust}. Whisper is an ASR system trained on a large and diverse dataset, enabling transcription in multiple languages and translation into English. WhisperX adds three components to Whisper. It is a practical choice for ASR in contexts where manual transcription is challenging to obtain. We applied WhisperX on the raw audio of the corpus \textit{Aug.} to have the automatic transcription and called the corpus \textit{Auto.} for \textit{Automatic transcription}.

\begin{table}[!ht]
    \centering
    \begin{tabular}{lrrrr}
        \toprule
        & \textbf{Hum.} & \textbf{Aug.} & \textbf{Test} \\
        \midrule
        \#dialog   & 200 & 1390 & 200 \\
        \#conv.len & 545 &  470 & 496 \\
        \#sum.lm   & 55.3 & 47.9 & 52.7 \\
        \bottomrule
    \end{tabular}
    \caption{DECODA distribution in the train (Hum. and Aug.) and the test set. \#dialog: number of dialogs; \#conv.len and \#sum.len average conversation and summary words respectively.}
    \label{tab:dist_corpus-DECODA-2}
\end{table}

Table \ref{tab:dist_corpus-DECODA-2} shows statistics for the training and test sets. Synopses in the test set are generally longer than in the training set. This disparity arises because the augmented data does not consistently mimic the style of human summaries in DECODA, given its 1-shot approach. The different datasets will be made available for research purposes on task-oriented dialog and particularly on task-oriented dialog summarization.

\subsection{Summarization and classification models}
\paragraph{Automatic Summarization} 
We trained the summarization systems using BARThez~\cite{kamal-eddine-etal-2021-barthez}, a sequence-to-sequence model pre-trained on various French corpora introduced for automatic text summarization. 
It is a transformer-based model built on BART architecture~\cite{vaswani2017attention, lewis-etal-2020-bart}. It consists of 6 encoder and decoder layers. 
We used the pre-trained model provided by the authors available on the Hugging Face library\footnote{\url{https://huggingface.co/moussaKam/barthez}} to train the models.
\paragraph{Call type classification}
Two classifiers based on CamemBERT-base \cite{martin-etal-2020-camembert} were trained to classify the call-type: one taking as input automatic dialog transcripts (Conv-CT-classifier) and the other one, generated synopses (Syn-CT-classifier). Conv-CT-classifier is also used to predict the call type for the $\text{Pipeline}_C$ model.

\subsection{Data augmentation and ASR evaluation}
We compute ROUGE, BERTScore, and the semantic score of text summarization systems trained on the different corpus versions to validate our data augmentation strategy using LLMs and ASR systems. We report all results in table \ref{tab:impact-data-aug}. For automatic transcription, we report result for two different sizes of WhisperX (Tiny-T and Large-L) and ChatGPT-3.5 one short on WhisperX Large transcription.
The findings indicate that data augmentation enhances summary scores while semantic scores remain constant, with a slight increase in NE-F1. In terms of automatic transcription, several metrics declined. However, the scores of the large model continue to be competitive, suggesting that the ASR system could be a viable option. ChatGPT-3.5 achieves the highest semantic scores in both manual and automatic transcriptions. This is unsurprising, considering the model's numerous parameters and the extensive training data it has received. In next section, only DECODA-v2-auto with WhisperX large transcription will be used for experiments.

\begin{table}[!ht]
    \centering
    \begin{tabular}{p{2cm}lrrrr}
        \toprule
        \textbf{Transcription} & \textbf{Data} & \textbf{ROUGE-L} & \textbf{BERTScore} & \textbf{CT-Acc} & \textbf{NE-F1} \\
        \midrule
        \multirow{3}{=}{\textbf{Manual transcription}} & 
         Hum. & 23.6 & 33.1 & 0.77 & 0.49\\
        & Aug. &29.1 & 38.9 & 0.77 & 0.51\\
        & one-shot*  & 28.9 & 37.3 & 0.80 & 0.51\\
        \midrule
        \multirow{3}{=}{\textbf{Automatic transcription}} & 
        Auto. (T) & 24.5 & 34.4 & 0.75 & 0.31\\
        & Auto. (L) & 27.7 & 37.1 & 0.76 & 0.43\\
        & one-shot* & 26.6 & 35.4 & 0.79 & 0.46\\
        \bottomrule
    \end{tabular}
    \caption{Automatic evaluation on the test partition with Barthez fine-tuned on the different datasets (Hum.,  Aug. Auto.). Performance of chatGPT3.5 with *. WER score of WhisperX Tiny (T) and Large (L) are respectively 76.4, 40.0.}
    \label{tab:impact-data-aug}
\end{table}

\subsection{Impact of selection criterion semantic criteria}
We trained $\text{Pipeline}_C$, the model that conditions the generation of the summary by the call type concatenated to the dialog using a separtor.
From this model, we generated various summaries using sampling decoding strategies and select the one according to the selection metrics presented in section \ref{subsec:summarySelection}. We report automatic results in table \ref{tab:selection} and denote as $\min D_{\text{KL}}$ for summaries selected using $D_{\text{KL}}$  and $\min \text{NEHR}$ for summaries selected using NEHR and $\min \text{NEHR} + D_{\text{KL}}$ for the combined criterion. We report the result for $\text{Pipeline}_C$ model and the baseline model without selection. For $\text{Pipeline}_C$, we have similar scores while CT-Acc vary. We can see that $D_{\text{KL}}$-based selection increases CT-Acc, while NEHR-based selection increases both values. The combined criteria seems to have the same result than $D_{\text{KL}}$-based selection.

\begin{table}[!ht]
    \centering
    \begin{tabular}{lrrrr}
        \toprule
        \textbf{System} & \textbf{ROUGE-L} & \textbf{BERTScore} & \textbf{CT-Acc} &  \textbf{NE-F1} \\
        \midrule
        Baseline & 27.7 & 37.1 & 0.76 & 0.43 \\
        $\text{Pipeline}_C$ & 27.6 & 37.0 & 0.79 & 0.43\\
        \midrule
        $\min \text{NEHR}$ & 27.7 & 36.4 & 0.81 & \textbf{0.46} \\
        $\min D_{\text{KL}}$ & 27.8 & 36.8 & \textbf{0.82}  & 0.44 \\
        $\min  \text{NEHR} + D_{\text{KL}}$ & 27.9 & 37.0 & \textbf{0.82} &  0.44 \\
        \bottomrule
    \end{tabular}
    \caption{BARThez + WhisperX Large + $\text{Pipeline}_C$ - Summaries evaluation of summary selection method.}
    \label{tab:selection}
\end{table}

\section{Discussion and Conclusion}
We analyzed the influence of task-semantic information on the generation of dialog summaries. Our findings revealed that incorporating NEHR in this model enhances named entity precision. Nonetheless, manual evaluation is necessary to validate the effectiveness of this selection criterion in terms of fidelity and informativeness. In addition,
we proposed new version of a dataset with more annotated data for task-oriented dialog summarization. Evaluation of the dataset revealed that using LLM to generate a portion of training synopses improved the model's performance, as indicated by increased ROUGE score and BERTScore, and automatic transcriptions can serve as a viable alternative to address the lack of annotated data.

\section*{Acknowledgements}
This work was granted access to the HPC resources of IDRIS under the allocation 2023-AD011012525R3 made by GENCI.

\appendix

\section{Details about DECODA dataset}
\label{sec:DECODA}
\begin{table}[!ht]
    \centering
    \begin{tabular}{lccccc}
        \toprule
        & DECODA-1 & DECODA-2 & DECODA-3\\
        \midrule
         Style & Synthetic & $\times$ & Literary\\
        \#dialog & 1009 & 478 & 500\\
        \#turn & 54.6 & 71.5 & 78.1\\
        \#conv.len & 411.8 & 513.3 & 504.1\\
        \#sum.len & 24.7 & $\times$ & 53.8\\
         \bottomrule
    \end{tabular}
    \caption{Description of DECODA corpus. \#dialog: total number of dialogs; \#turn: average number of turns; \#conv.len and \#sum.len average dialog and summary number of words respectively. $\times$ No summary available.}
    \label{tab:DECODA-corpus-part}
\end{table}

\section{Prompt used to generate summary}
\label{sec:prompt}

To generate summary from transcription using ChatGPT-3.5, we used the prompt (translated from French) and one-shot example in table \ref{tab:prompt-one-shot-example}.
\begin{table}[!ht]
    \centering
    \begin{tabular}{p{2cm}|p{13cm}}
        \toprule
        \multicolumn{2}{c}{\textbf{Prompt translated from French}}\\
        \midrule
        \textbf{Prompt} & \textit{Summarize concisely in correct French, without giving too many details, the following text which represents a dialogue between \#APPELANT who uses Parisian transport and \#CONSEILLER of the RATP. Each turn begins with the name of the speaker in square brackets, followed by the transcription of the turn. Each turn ends with the <END> symbol. Here's the transcript to summarize:} \\
        \midrule
        \multicolumn{2}{c}{\textbf{One-shot example (translated from French)}} \\
        \midrule
        \textbf{Dialog} & [null] will answer you <END> [agent] good morning <END> [customer] yes good morning madam <END> [customer] madam I'd like to know if tomorrow morning there will be any disruption on the B line of the RER towards Roissy - Charles-de-Gaulle <END> [agent] so the disruptions on the RER B are only in the evening at the airport but not during the day <END> [customer] ah ah are there any disruptions planned for tomorrow evening ? <END> [agent] uh yes because you have works on the section <END> [agent] SNCF so <END> [customer] ah well without that <END> [customer] it works normally? <END> [agent] so if you want that's why but if you take you you leave at what time? <END> [agent] in the morning? <END> [customer] oh <END> [customer] very early <END> [agent] then in that case no <END> [customer] uh ve around 6 o'clock in the <END> [customer] morning <END> [agent] no no there's no problem <END> [customer] there's no problem <END> [agent] no no no no <END> [customer] it's on E only that there are disruptions? <END> [agent] then on the E yes you have d yes yes quite where there it's uh it's also under construction <END> [agent] there's a lot of <END> [customer] <END> [agent] work currently yes <END> [customer] good! in other words uh p tomorrow oh uh no problem on the B <END> [agent] hm no <END> [customer] thank you very much <END> [agent] but I'll <END> [agent] please goodbye sir <END> [customer] goodbye madam <END>. \\
        \hline
         \textbf{Summary} &  A caller wishes to know whether RER Line B will be disrupted the following day, due to work on the route. The advisor says that disruption is expected the next day, but only in the evening.\\
         \bottomrule
    \end{tabular}
    \caption{Prompt and One-shot example for ChatGPT 3.5}
    \label{tab:prompt-one-shot-example}
\end{table}

\section{Call type classification systems}
We report the call type classifier used to generate the call type of the $\text{Pipeline}_C$ system and to perform the $D_{\text{KL}}$-based selection in table \ref{tab:classificationModel-2}
\begin{table}[!ht]
    \centering
    \begin{tabular}{lrrr}
        \toprule
        \textbf{System} &  \textbf{Acc.} & \textbf{W-F1} \\
        \midrule
        Conv-CT-classifier & 81 & 80 \\
        Syn-CT-classifier & 80 & 80 \\
        \bottomrule
    \end{tabular}
    \caption{Accuracy and Weighted F1 Score of call types classifiers}
    \label{tab:classificationModel-2}
\end{table}

\section{All WhisperX size results}
We report in the table \ref{tab:wer-full} all the results obtained by training text summarization system on different whisperX size. 

\begin{table}[!ht]
    \centering
    \begin{tabular}{lrrrrrrrr}
        \toprule
        \textbf{Transcription} & \textbf{WER} & \textbf{ROUGE-L} & \textbf{BERTScore} & \textbf{CT-Acc} & \textbf{NE-P} & \textbf{NE-R}\\
        \midrule
        Manual & \textbf{0.0} & \textbf{29.1} & \textbf{38.9} & 0.77 & \textbf{0.54} & \textbf{0.48} \\
        WhisperX Tiny & 76.4 & 24.5 & 34.4 & 0.75 & 0.35 & 0.28 \\
        WhisperX Base & 58.6 & 26.2 & 35.9 & 0.75 & 0.39 & 0.30 \\
        WhisperX Small & 46.4 & 27.0 & 36.9 & 0.75 & 0.49 & 0.35 \\
        WhisperX Medium & 40.9 & 27.6 & 36.8 & \textbf{0.79} & 0.51 & 0.35 \\
        WhisperX Large & 40.0 & 27.7 & 37.1 & 0.76 & 0.53 & 0.36 \\
        \bottomrule
    \end{tabular}
    \caption{WER score on automatic transcription from different size of WhisperX.
    ROUGE-L and BERT-scores on summaries generated by Barthez finetuned and evaluated on automatic transcriptions. CT-Acc, NE-P, NE-R, and NE-F1 are our semantic measures based on call type and named entities.}
    \label{tab:wer-full}
\end{table}

\section{Sampling parameters}
\begin{itemize}
    \item Top-P: $P \in [0.70, 0.95[$ with the step of 0.05
    \item Top-K: $K \in [30, 100[$ with the step of 15
    \item Temperature: $T \in [0.7, 1]$ with the step of 0.1
    \item Greedy decoding
    \item Beam decoding: We used a beam of size 6 and keep the 6-best.
\end{itemize}

\section{Selection results}
We report in table \ref{tab:selection-all} all the scores obtained for summary selection. We include here, the result obtained when using the whisperX base size as automatic transcription system. 
\begin{table*}[!ht]
    \centering
    \begin{tabular}{p{2cm}lrrrrrr}
        \toprule
        \textbf{System} & \textbf{Criteria} & \textbf{ROUGE-L} & \textbf{BERTScore} & \textbf{CT-Acc} &  \textbf{NE-P} & \textbf{NE-R} & \textbf{NE-F1}\\
        \midrule
        \multirow{4}{=}{BARThez + $\text{Pipeline}_C$}
        & \textsc{Beam} = 4 & 27.6 & 37.0 & 0.79 & 0.54 & 0.36 & 0.43 \\
        & $\min \text{NEHR}$ & 27.7 & 36.4 & 0.81 & \textbf{0.58} & 0.38 & \textbf{0.46}\\
        & $\min D_{\text{KL}}$ & 27.8 & 36.8 & \textbf{0.82}  & 0.52 & 0.38 & 0.44\\
        & $\min \text{NEHR} + D_{\text{KL}}$ & 27.9 & 37.0 & \textbf{0.82} &  0.51 & \textbf{0.39} & 0.44 \\
        \midrule
        \multirow{4}{=}{BARThez + WhisperX Base}
        & \textsc{Beam} = 4 & 26.2 & 35.9 & 0.75 & 0.39 & 0.30 & 0.34 \\
        & $\min \text{NEHR}$ & 25.7 & 35.5 & 0.75  & 0.45 & 0.40 & 0.42 \\
        & $\min D_{\text{KL}}$ & 24.8 & 33.0 & \textbf{0.79} &  0.40 & 0.31 & 0.35 \\
        & $\min \text{NEHR} + D_{\text{KL}}$ & 25.5 & 35.5 & 0.77 & \textbf{0.46} & \textbf{0.43} & \textbf{0.44}\\
        \bottomrule
    \end{tabular}
    \caption{Evaluation of summary selection method. For the model BARThez + $\text{Pipeline}_C$ the dataset: DECODA-v2-auto with Whisper Large. For BARThez + WhisperX Base, the dataset DECODA-v2-auto with transcription from WhisperX Base }
    \label{tab:selection-all}
\end{table*}

\bibliographystyle{unsrtnat}
\bibliography{references}  

\end{document}